\documentclass{article}
\usepackage{spconf,amsmath,graphicx}

\usepackage[colorlinks, 
linkcolor=red, anchorcolor=blue, citecolor=green, urlcolor=blue
]{hyperref,xcolor}
\usepackage{makecell}
\usepackage{url}
\usepackage{wrapfig}
\usepackage{dsfont}
\usepackage{subfigure}
\usepackage{booktabs} 
\usepackage{latexsym}
\usepackage{threeparttable}
\usepackage{multirow}
\usepackage{amsfonts}
\usepackage{bbm}
\usepackage{bm}
\usepackage{bbold}
\usepackage{color}
\usepackage{enumitem}
\usepackage{mathalfa}
\usepackage{bbding}

\newcommand{\bluebold}[1]{\textcolor{blue}{\textbf{#1}}}

\title{Communication-Efficient Personalized Federated Learning for Speech-to-Text Tasks}
\name{
Yichao Du$^{\ddag }$\thanks{$^*$Work done during Yichao's internship at Tencent AI Lab. }, Zhirui Zhang$^{\dag}$, Linan Yue$^{\ddag}$, Xu Huang$^{\natural}$, Yuqing Zhang$^{\flat}$, Tong Xu$^{\ddag}$, Linli Xu$^{\ddag}$, Enhong Chen$^{\ddag}$
}

\address{
$^{\ddag}$University of Science and Technology of China \& State Key Laboratory of Cognitive Intelligence 
\\ $^{\dag}$Tencent AI Lab  $^{\natural}$Nanjing University $^{\flat}$Anhui Xinhua University
}
\begin{document}

\maketitle
\begin{abstract}
To protect privacy and meet legal regulations, federated learning (FL) has gained significant attention for training speech-to-text (S2T) systems, including automatic speech recognition (ASR) and speech translation (ST). 
However, the commonly used FL approach (i.e., \textsc{FedAvg}) in S2T tasks typically suffers from extensive communication overhead due to multi-round interactions based on the whole model and performance degradation caused by data heterogeneity among clients.
To address these issues, we propose a personalized federated S2T framework that introduces \textsc{FedLoRA}, a lightweight LoRA module for client-side tuning and interaction with the server to minimize communication overhead, and \textsc{FedMem}, a global model equipped with a $k$-nearest-neighbor ($k$NN) classifier that captures client-specific distributional shifts to achieve personalization and overcome data heterogeneity. 
Extensive experiments based on Conformer and Whisper backbone models on CoVoST and GigaSpeech benchmarks show that our approach significantly reduces the communication overhead on all S2T tasks and effectively personalizes the global model to overcome data heterogeneity.

\end{abstract}

\section{Introduction}
\label{sec:intro}

Industrial-grade speech-to-text (S2T) systems, including automatic speech recognition (ASR) and end-to-end speech translation (ST), require massive speech data to achieve high performance, typically involving the collection of client data for centralized training. However, speech data inherently possess privacy-sensitive characteristics~\cite{Schuller2013Computational} and often exist in isolated silos. For commercial considerations and government regulations (e.g., GDPR and CCPA), the aforementioned training paradigm is not always applicable. 
Recently, researchers have introduced federated learning (FL) into ASR task~\cite{Guliani2020TrainingSR,Gao2022EndtoEndSR,Cui2021FederatedAM,Nguyen2023FederatedLF,Jia2022FederatedDA,Mehmood2022FedNSTFN,Yang2020DecentralizingFE,Rao2023FederatedSW,Zhu2022DecoupledFL}, enabling multiple clients to collaboratively train a global model under the coordination of a service provider while keeping their private data localized. Despite the flourishing research in federated ASR, the study of federated speech translation remains unexplored. In this paper, we aim to investigate both ASR and ST under a unified federated S2T perspective.

Existing federated S2T workflows primarily follow the \textsc{FedAvg}~\cite{McMahan2016CommunicationEfficientLO} paradigm, which involves multi-round whole-model interactions between the server and clients. In practice, training S2T systems in this manner faces numerous challenges. On one hand, state-of-the-art S2T models typically rely on large backbones (e.g., Transformer~\cite{Vaswani2017AttentionIA,Du2021RegularizingES,Radford2022RobustSR} and Conformer~\cite{Gulati2020ConformerCT,Du2023TheMS}) to handle high-dimensional acoustic features, resulting in substantial computational and communication overhead when applying FedAvg directly. 
Moreover, clients exhibit heterogeneous hardware capabilities, making it difficult to accommodate stringent requirements for whole-model training and communication.
On the other hand, speech data distribution exhibits severe heterogeneity~\cite{Guliani2020TrainingSR,Gao2022EndtoEndSR,Zhu2022DecoupledFL} (e.g., variations in environmental noise, accents, domains, microphone conditions), which can lead to parameter conflicts among client models. 
Simply employing the FedAvg algorithm to aggregate parameters may result in an insufficiently personalized global model, leading to a significant decline in performance. Therefore, it remains a crucial issue: \textit{\textbf{How to efficiently train federated S2T systems and personalize them?}}

\begin{figure*}[h]
    \small
    \centering
    \includegraphics[width = 17cm]{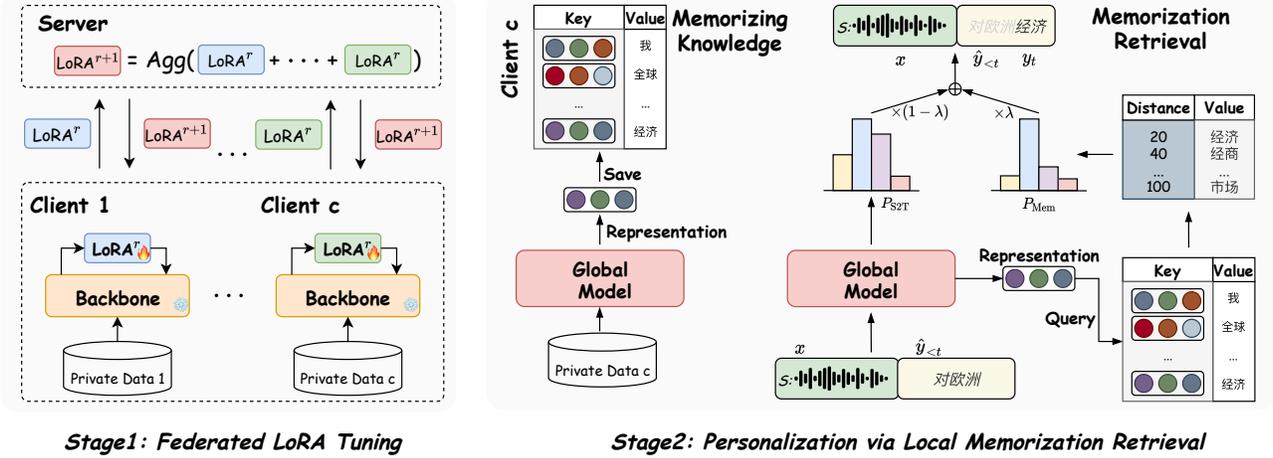}
    \caption{An overview of our proposed efficient personalized federated S2T framework.}
    \label{fig:framework}
\end{figure*}

To tackle the aforementioned issues, we propose the \textit{first} efficient personalized federated S2T framework, which is decoupled into two stages: 
(i) In the first stage, we introduce the \textsc{FedLoRA} to train a global model. In each round, each client freezes the backbone model's parameters and adopts a lightweight Low-Rank Adaptation (LoRA)~\cite{Hu2021LoRALA} module for parameter-efficient tuning and interacts with the server. In this way, we can easily satisfy communication and computational constraints.
(ii) In the second stage, we introduce \textsc{FedMem} to equip the well-trained global model with a $k$-nearest-neighbor ($k$NN) classifier that does not require additional training. This classifier captures local distribution shifts relative to the global distribution by utilizing client-specific memorization retrieval, thereby achieving personalization and overcoming the clients' model parameter conflicts caused by data heterogeneity. 

We evaluate our proposed framework on the dialect benchmark CoVoST~\cite{Wang2021CoVoST2A} and the multi-domain benchmark GigaSpeech~\cite{Chen2021GigaSpeechAE}. 
Experimental results demonstrate that the \textsc{FedLoRA} can reduce communication overhead by up to 96.5$\%$ while maintaining comparable or even superior performance to the centralized model. Furthermore, the \textsc{FedMem} leads to improving all global models, substantiating the effectiveness of memorization retrieval for personalization to mitigate data heterogeneity.

\section{Methods}
\subsection{Federated Speech-to-Text Tasks}
In this work, we focus on a generalized application scenario, i.e., \textit{cross-silo} FL, to investigate ASR and ST under the unified federated S2T perspective. 
Formally, this scenario consists of $|C|$ clients and a central server.  
Following the previous work~\cite{Gao2022EndtoEndSR,Zhu2022DecoupledFL}, the central server provides a pre-trained backbone S2T model $f_{\theta^0}$, where $\theta^0$ denotes model parameters. For each client $c$, it contains private data $\mathcal{D}_c=\{(\mathbf{x}_c^{i}, \mathbf{y}_c^{i})\}_{i=1}^{|\mathcal{D}_c|}$ and only accessible to itself, where $\mathbf{x}^{i}$ 
and $\mathbf{y}^{i}$ are sequences in the source speech and target text, respectively.
The straightforward training idea is to apply the \textsc{FedAvg}.
At each round $r$, each client $c$ downloads $f_{\theta^r}$ from the server and optimizes it on $\mathcal{D}_c$ via maximum likelihood estimation:
\begin{equation}
\mathcal{L}_{S2T}(\theta^r_c)=\sum\nolimits_{i=1}^{|\mathcal{D}_c|} \log P_{\textrm{S2T}}\left(\mathbf{y}_c^i \mid \mathbf{x}^i_c ; \theta_c^r\right)
\end{equation}
Then the local updates $\theta^{r}_c$ are uploaded to the server, while the server aggregates them to form a new model $f_{\theta^{r+1}}$ via a simple parameter averaging: $\theta^{r+1} = \sum_{m=1}^{C} \frac{n_m}{n} \ \theta^{r}_m$, where $n_m$ denotes the number of private data points in the $m$-th client’s, and $n$ is the total number of all client data. However, the large model size and the system heterogeneity among clients result in low communication efficiency. Additionally, the data heterogeneity leads to a lack of personalization and a subsequent decline in performance.

\subsection{Efficient Personalized Federated S2T Framework}
To cope with the above issues, we propose an efficient personalized federated S2T framework, which is decoupled into two stages: 1) \textsc{\textbf{FedLoRA}} for efficient training of global model and 2) \textsc{\textbf{FedMem}} for personalization based on client-specific memorization retrieval.

\subsubsection{Stage 1: Federated LoRA Tuning}
To efficiently train the global model, we design \textsc{FedLoRA}, which employs LoRA~\cite{Hu2021LoRALA} as the object to be interacted with and updated. 
The LoRA tuning technique injects trainable rank decomposition matrices into each layer of the backbone model and freezes the original weights. Its parameter update can be formalized as: 
\begin{equation}
    W_0+\Delta W=W_0+B A
    \label{eq:lora}
\end{equation}
where $W_0$ denotes the original weights in the backbone model $f_{\theta^0}$. $\Delta W=BA$ denotes the updates of the LoRA module, where $B$ and $A$ are two low-rank matrices.
During the \textsc{FedLoRA} training, at each round $r$, client $c$ first optimizes the local LoRA module $f_{\delta}$ using the following loss:
\begin{equation}
    \mathcal{L}_{S2T}(\delta^{r}_{c})=\sum\nolimits_{i=1}^{|\mathcal{D}_c|} \log P_{\textrm{S2T}}\left(\mathbf{y}_c^i \mid \mathbf{x}^i_c ; {\theta^0}, \delta^{r}_{c}\right)
\end{equation}
Then the local updates $\delta^{r}_{c}$ are uploaded to the server, while the server aggregates these updates to form a new LoRA module $f_{\delta^{r+1}}$ via a simple parameter averaging technique: $\delta^{r+1} = \sum_{m=1}^{C} \frac{n_m}{n} \ \delta^{r}_m$. 
In the last round, according to Equation~\ref{eq:lora}, each client directly integrates the well-trained global LoRA module $f_{\delta}$ into the backbone model to obtain the final global S2T model $f_{\theta_g}$ (where $\theta_g = {\theta^0} + \delta$). In this way, $f_{\theta_g}$ can update the whole backbone with tiny overhead and maintain the original inference speed. 

\subsubsection{Stage 2: Personalization via Memorization Retrieval}
Inspired by the recent success of $k$NN memorization retrieval techniques in natural language processing~\cite{Khandelwal2019GeneralizationTM,Khandelwal2020NearestNM,Wang2021NonParametricOL,Marfoq2021PersonalizedFL,Du2022NonParametricDA,Nie2022ImprovingFP,Du2023FederatedNN,Gao2023NearestNM}, we propose the \textsc{FedMem} strategy. This strategy incorporates a $k$NN classifier into a well-trained global model, allowing for client-specific memorization retrieval that capture the local distribution shifts relative to the global one.
This way facilitates personalization and then alleviates conflicts arising from data heterogeneity in global model's parameters.
It mainly involves two steps:

\par\noindent\ignorespacesafterend 
\textbf{Memorizing Client-Specific Knowledge into Datastore.}
For each client $c$, the local datastore is the cache of a set of key-value pairs. 
After obtaining the global model $f_{\theta_g}$, at each time-step $t$, we use teacher-forcing mode to obtain the context representation $f_{\theta_g} (\mathbf{x}, y_{\textless t})$ on the parallel sentence pair $(\mathbf{x},\mathbf{y}) \in \mathcal{D}_c$.
Then the whole datastore $(\mathcal{K},\mathcal{V})$ is constructed by taking the $f_{\theta_g}(\mathbf{x}, y_{\textless t})$ as key and ground-truth $y_t$ as value:
\begin{align}
(\mathcal{K}, \mathcal{V}) = \bigcup_{(\mathbf{x}, \mathbf{y}) \in \mathcal{D}_c} \{(f_{\theta_g}(\mathbf{x}, y_{\textless t}), y_t), \forall y_t \in \mathbf{y} \}. 
\end{align}

\par\noindent\ignorespacesafterend 
\textbf{Inference with Memorized Knowledge.}
During inference, at time step $t$, \textsc{FedMem} leverage the context representation $f_{\theta_g}(\mathbf{x},\hat{y}_{\textless t})$ to query the cached datastore $(\mathcal{K}, \mathcal{V})$ and retrieves $k$ nearest neighbor key-value pairs w.r.t. Euclidean distance. Then the probability distribution of $y_t$ generated by $k$NN-based memorization retrieval is calculated as follows:
\begin{align}
    p_{\textrm{Mem}}&(y_t|\mathbf{x}, \hat{y}_{\textless t}) \propto \label{eq:knn_prob} \sum_{(h_i, v_i)\in \mathcal{R}} \mathbb{1}_{y_t = v_i} \exp (\frac{-d(h_i, f_{\theta_g}(\mathbf{x}, \hat{y}_{<t}))}{T}), \nonumber
\end{align}
where $\mathcal{R} = \{(h_i, v_i), i \in \{1, 2, ..., k\} \}$ is the set of $k$ nearest neighbors, $d(\cdot , \cdot)$ represents the squared Euclidean distance and $T$ is the temperature to control the sharpness of softmax function. 
The final output distribution is an interpolation between distributions from the $f_{\theta_g}$ and memorization retrieved neighbors with a interpolation coefficient $\lambda \in [0, 1]$:
\begin{equation}
\begin{split}
    p(y_t|\mathbf{x},\hat{y}_{\textless t}) & = \lambda \ p_{\textrm{Mem}}(y_t|\mathbf{x},\hat{y}_{\textless t}) \\
& + (1-\lambda) \ p_{\textrm{S2T}}(y_t|\mathbf{x},\hat{y}_{\textless t}). 
     \end{split}
\label{eq:final_prob}
\end{equation}
Note that, this prediction way could also be substituted with other memorization-retrieval variants~\cite{Zheng2021AdaptiveNN,Meng2021FastNN,Martins2022ChunkbasedNN,Dai2023SimpleAS} to achieve better performance or inference speed, we leave it as future work. 

\begin{table*}[htb]
    \small
    \centering
    \caption{The ASR performance of different methods on CoVoST. ``$\text{\#}$Params" and ``$\text{\#}$Comm" refer to the number of parameters and communication overhead, respectively.}
    \label{tab:covost_asr_res}
    \setlength{\tabcolsep}{1.2mm}{
    \scalebox{0.845}{
        \begin{tabular}{l|r|ccccc|cc|r|ccccc|ccccccccccccc}
        \toprule
        \multicolumn{1}{l|}{\multirow{2}{*}{\textbf{Methods}}}
        & \multicolumn{8}{c|}{\textbf{Conformer Backbone ($\text{\#}$Params, WER, $\text{\#}$Comm})} & \multicolumn{8}{c}{\textbf{Whisper Backbone ($\text{\#}$Params, WER, $\text{\#}$Comm})} \\
        
        & \multicolumn{1}{c}{\textbf{Tuned/Total}} & \textbf{AU} & \textbf{CA} & \textbf{EN} & \multicolumn{1}{c}{\textbf{IN}} & \multicolumn{1}{c}{\textbf{Avg}}  & \textbf{Round} & \textbf{Cost} & \multicolumn{1}{c}{\textbf{Tuned/Total}} & \textbf{AU} & \textbf{CA} & \textbf{EN} & \multicolumn{1}{c}{\textbf{IN}} & \multicolumn{1}{c}{\textbf{Avg}} & \textbf{Round} & \textbf{Cost}\\
        \midrule
        \textsc{P-S2T}                  & 140M/140M   & 62.87  & 53.65  & 62.22  & 72.72  & 62.87  & -- & -- & 244M/244M   & 15.75  & 12.28  & 13.47  & 21.10  & 15.65  & -- & -- \\ 
        \midrule
        \textsc{C-S2T}$_\textsc{Full}$  & 140M/140M   & \bluebold{20.05}  & \bluebold{16.30}  & \bluebold{18.40}  & \bluebold{28.14}  & \bluebold{20.72} & -- & --       & 244M/244M   & 15.27  & 12.88  & 14.07  & 17.03  & 14.81 & -- & --  \\
        \textsc{FedAvg}                 & 140M/140M & 21.83  & 17.51  & 19.99  & 32.08  & 22.85 & 82 & 344.22GB & 244M/244M   & 15.40  & 13.19  & 14.14  & 16.80  & 14.88  & 15 & 112.71 GB \\
        + \textsc{FedMem}               & 0M/140M     & 21.53  & 17.19  & 19.73  & 31.67  & 22.53 & -- & --       & 0M/244M     & 15.11  & 12.84  & 13.89  & \bluebold{16.47}  & 14.58  & -- & --   \\
        \midrule
        \textsc{C-S2T}$_\textsc{LoRA}$  & 4.5M/144.5M   & 22.52  & 17.91  & 20.51  & 31.59  & 23.13 & -- & --      & 10.1M/254.1M   & 14.81  & 12.57  & 13.83  & 16.73  & 14.48 & -- & --      \\
        \textsc{FedLoRA}                & 4.5M/144.5M & 23.34  & 18.58  & 22.32  & 32.97  & 24.30 & 74 & \bluebold{12.01GB} & 10.1M/254.1M & 14.78  & 12.40  & 13.57  & 16.88  & 14.41 & 20 & \bluebold{9.66 GB} \\
        + \textsc{FedMem}               & 0M/140M     & 23.10  & 18.02  & 22.11  & 31.76  & 23.75 & -- & --      & 0M/244M     & \bluebold{14.52}  & \bluebold{12.19}  & \bluebold{13.42}  & 16.67  & \bluebold{14.20} & -- & --      \\
        \bottomrule
    \end{tabular}
    }}
\end{table*}

\section{Experiments}
\subsection{Setup}
We adopt two federated S2T scenarios to conduct experiments, in which natural data exists heterogeneity: i) \textbf{dialect} setting, which uses four accents from the \textbf{CoVoST}~\cite{Wang2021CoVoST2A} benchmarks (English ASR and English-Chinese ST) as the four clients. ii) \textbf{multi-domain} setting, which uses ten domains from the \textbf{GigaSpeech}~\cite{Chen2021GigaSpeechAE} benchmark (English ASR) as the clients involved in FL training. 
The additional six domains as invisible clients to investigate the test-time generalization ability. 
Table~\ref{tab:data_stat} details the statistics of these datasets.

\begin{table}[htb]
    \small
    \centering
    \renewcommand\arraystretch{0.88}
    \caption{The statistics of MuST-C, CoVoST and GigaSpeech. 
    The `visiable" indicate that client whether involved in global model training.
    }
    \label{tab:data_stat}
    \setlength{\tabcolsep}{1.25mm}{
    \scalebox{1.0}{
        \begin{tabular}{c|l|cccccccc} 
            \toprule
            &  & \textbf{Hours}    & \textbf{Train} & \textbf{Dev}  & \textbf{Test}   & \multirow{1}{*}{\textbf{Visible}} \\
            \midrule
            & \multirow{1}{*}{\textbf{MuST-C}}   & 983 & 531,675 & 2,210 & 4,557  & -- \\
            \midrule
            \multirow{4}{*}{\rotatebox[]{90}{\textbf{CoVoST}}}
            & \textbf{England (EN)}     & 28 & 19,188 & 2,500 & 2,500  & \checkmark \\
            & \textbf{Australia (AU)}     & 19 & 12,209 & 2,500 & 2,500  & \checkmark \\
            & \textbf{Indian (IN)}     & 18 & 11,396 & 2,500 & 2,500  & \checkmark \\
            & \textbf{Canada (CA)}     & 14 & 9,151  & 2,500 & 2,500  & \checkmark \\
            \midrule
            \multirow{17}{*}{\rotatebox[]{90}{\textbf{GigaSpeech}}} 
            & \textbf{Howto}      & 154  & 12,4290   & 3,039 & 3,938 & \checkmark \\
            & \textbf{Crime}      & 109  & 95,715    & 3,658 & 2,870 & \checkmark \\
            & \textbf{Health}     & 127  & 83,642    & 3,720 & 2,975 & \checkmark \\
            & \textbf{Film}       & 75   & 73,816    & 3,093 & 3,194 & \checkmark \\
            & \textbf{Gaming}     & 53   & 55,047    & 2,479 & 2,216 & \checkmark \\
            & \textbf{Arts}       & 71   & 48,146    & 2,131 & 2,688 & \checkmark \\
            & \textbf{Sports}     & 31   & 29,519    & 3,675 & 3,300 & \checkmark \\
            & \textbf{Comedy}     & 30   & 28,370    & 1,726 & 2,435 & \checkmark \\
            & \textbf{History}    & 33   & 23,835    & 3,299 & 3,294 & \checkmark \\
            & \textbf{Society}    & 27   & 21,377    & 2,634 & 2,474 & \checkmark \\
            \cmidrule[.5pt]{2 - 7}
            & \textbf{News}             & 2827 & 1,888,342 & 3,159 & 3,162 & \XSolidBrush \\
            & \textbf{Education}        & 1663 & 1,253,807 & 3,268 & 5,494 & \XSolidBrush \\
            & \textbf{Sci. \& Tech}             & 715  & 48,8051   & 2,831 & 2,550 & \XSolidBrush \\
            & \textbf{People}           & 524  & 416,908   & 3,021 & 1,760 & \XSolidBrush \\
            & \textbf{Entertainment}    & 288  & 264,214   & 2,548 & 3,879 & \XSolidBrush \\
            & \textbf{Nonprofits}       & 188  & 149,343   & 2,953 & 2,319 & \XSolidBrush \\
            \bottomrule
        \end{tabular}
    }}
\end{table}

\begin{table}[htb]
    \small
    \centering
    \caption{The ST performance on CoVoST.}
    \label{tab:covost_st_res}
    \setlength{\tabcolsep}{1.0mm}{
    \scalebox{0.78}{
        \begin{tabular}{l|r|ccccc|cccccccccc}
        \toprule
        \multicolumn{1}{l|}{\multirow{2}{*}{\textbf{Methods}}}
        & \multicolumn{8}{c}{\textbf{Conformer Backbone ($\text{\#}$Params, BLEU, $\text{\#}$Comm})} \\
        & \multicolumn{1}{c}{\textbf{Tuned/Total}} & \textbf{AU} & \textbf{CA} & \textbf{EN} & \multicolumn{1}{c}{\textbf{IN}} & \multicolumn{1}{c}{\textbf{Avg}}  & \textbf{Round} & \textbf{Cost}\\
        \midrule
        \textsc{P-S2T}                  & 146M/146M  & 15.59  & 18.96  & 17.46  & 11.43  & 15.86 & -- & -- \\ 
        \midrule
        \textsc{C-S2T}$_\textsc{Full}$  & 146M/146M   & \bluebold{27.78}  & \bluebold{29.77}  & \bluebold{27.84}  & \bluebold{23.04}  & \bluebold{27.11}  & -- & --\\
        \textsc{FedAvg}                 & 146M/146M   & 26.60  & 29.31  & 27.47  & 21.03  & 26.10  & 46 & 202.33GB\\
        + \textsc{FedMem}               & 0M/146M     & 27.13  & 29.83  & 27.71  & 21.24  & 26.48  & -- & -- \\
        \midrule
        \textsc{C-S2T}$_\textsc{LoRA}$  & 4.5M/151.5M & 27.66  & 29.27  & 27.54  & 21.73  & 26.55 & -- & --\\
        \textsc{FedLoRA}                & 4.5M/151.5M & 26.53  & 28.69  & 26.26  & 20.65  & 25.53 & 52 & \bluebold{9.15GB} \\
        + \textsc{FedMem}               & 0M/146M     & 26.91  & 29.12  & 26.92  & 21.24  & 26.05 & -- & --\\
        \bottomrule
    \end{tabular}
    }}
\end{table}

\subsubsection{Baselines} 
We compare \textsc{FedLoRA}/\textsc{FedMem} with several baselines: 
(i) \textbf{Public Model (\textsc{P-S2T})}: a pre-trained S2T model provided by the server and used to initialize the client's model. 
(ii) \textbf{Centralized Model (\textsc{C-S2T})}: a standard centralized training method that uses all clients' speech data to train a global S2T model. 
It contains two versions, i.e., full fine-tuning and LoRA tuning. 
(iii) \textbf{\textsc{FedAvg}}~\cite{McMahan2016CommunicationEfficientLO}: a vanilla FL approach which optimizes the global S2T model based on multi-round iterations with whole models.

\begin{table*}[htb]
    \small
    \centering
    \renewcommand\arraystretch{0.85}
    \caption{The ASR performance of different Whisper-based methods on GigaSpeech.}
    \label{tab:gigaspeech_asr_res}
    \setlength{\tabcolsep}{1.25mm}{
    \scalebox{0.91}{
    \begin{tabular}{l|r|cccccccccc|c|cccc}
        \toprule
        \multicolumn{1}{l|}{\multirow{2}{*}{\textbf{Methods}}}
        & \multicolumn{14}{c}{\textbf{Whisper Backbone ($\text{\#}$Params, WER, $\text{\#}$Comm})} \\
        & \multicolumn{1}{c}{\textbf{Tuned/Total}} & \textbf{Arts}  & \textbf{Comedy}  & \textbf{Crime}  & \textbf{Film}  & \textbf{Gaming}  & \textbf{Health}  & \textbf{History}  
        & \textbf{Howto}  & \textbf{Society}  & \multicolumn{1}{c}{\textbf{Sports}} & \multicolumn{1}{c}{\textbf{Avg}}  & \multicolumn{1}{c}{\textbf{Round}} & \textbf{Cost} \\
        \midrule
        \textsc{P-S2T}         & 244M/244M & 18.45  & 33.09  & 20.79  & 37.78  & 30.78  & 18.64  & 18.31  & 23.56  & 20.02  & 29.07  & 25.05 & -- & --  \\
        \midrule
        \textsc{C-S2T}$_\textsc{Full}$ & 244M/244M & 5.40  & 9.68  & 5.70  & 9.54  & 7.64  & 5.34  & 3.38  & 7.31  & 4.57  & 11.12   & 6.97  & --  & -- \\
        \textsc{FedAvg}                & 244M/244M & 5.44  & 9.56  & 5.89   & 8.67   & 7.74   & 5.42  & 8.20  & 7.23   & 4.56  & 11.32  & 7.40  & 13 & 245.42GB \\ 
        + \textsc{FedMem}              & 0M/244M    & 5.13  & 9.23  & 5.23   & 8.12   & 7.29   & 4.76  & 7.95  & 6.87   & 4.41  & 11.10  & 7.01  & --  & -- \\
        \midrule
        \textsc{C-S2T}$_\textsc{LoRA}$  & 10.1M/254.1M & 5.63  & 8.92  & 5.80  & 9.05  & 7.87  & 5.56  & 2.75  & 6.85  & 4.46  & 11.24  & 6.81 & --  & --     \\ 
        \textsc{FedLoRA}                & 10.1M/254.1M & 5.47  & 8.60  & 5.66  & 8.65  & 7.44  & 5.39  & 2.76  & 7.24  & 4.57  & 10.87  & 6.67 & 15 & \bluebold{20.38GB} \\  
        + \textsc{FedMem}               & 0M/244M      & \bluebold{5.10}  & \bluebold{8.19}  & \bluebold{5.17}  & \bluebold{8.32}  & \bluebold{6.78}  & \bluebold{4.79}  & \bluebold{2.69}  & \bluebold{6.75}  & \bluebold{4.50}  & \bluebold{10.52}  & \bluebold{6.28} & -  & -     \\
        \bottomrule
    \end{tabular}}}
\end{table*}

\subsubsection{Training Details and Evaluation.}
For the aforementioned settings, the clients utilize the pre-trained models (i.e., Conformer~\cite{Gulati2020ConformerCT} or Whisper~\cite{Radford2022RobustSR}) provided by the server for initialization. We pre-train Conformer on MuST-C~\cite{Gangi2019MuSTCAM} data in Table~\ref{tab:data_stat}. It consists of 12 encoders and 6 decoders, with the embedding size, FFN size, and heads being 512, 2048, and 4, respectively. Each clients adopt an Adam with a learning rate of 1e-4 and 500 warm-up updates. We follow the \textsc{Fairseq-S2T}\footnote{\url{https://github.com/facebookresearch/fairseq/blob/main/examples/speech_to_text}} 
recipe to process the speech data, and use \textsc{SentencePiece}\footnote{\url{https://github.com/google/sentencepiece}} 
to build 10K/16K unigram sub-words for the English/Chinese text.
We also employ the Whisper-small to conduct experiment, where 
each client's learning rate and warm-up steps are set to 5e-5 and 1K, respectively. We add the LoRA module to the $W_Q, W_K, W_V, W_{proj}, FC_1 \text{and}\ FC_2$ of the self-attention and set rank/alpha 32/64 respectively. All training is conducted on 8 V100 GPUs. In each round of interaction, the client update epoch is set to 1, and the global patience is set to 5 to select the optimal rounds. \textsc{FedMem} employs the \textsc{Faiss}\footnote{\url{https://github.com/facebookresearch/faiss}} 
library to build the datastore and quantizes it into 4096 clusters using IndexIVFPQ, which can reduce storage and accelerate inference. During inference, the beam size is set to 5. When applying FedMem, we search 64 clusters for each target token. We search for the best $k\in\{4,8,16\}, \lambda\in\{0.1,0.2,...,0.9\}\ \text{and}\ T\in\{10, 20, 50, 100, 200\}$ on the dev set. We evaluate ASR and ST using case-sensitive WER and SacreBLEU\footnote{\url{https://github.com/mjpost/sacrebleu}}
, respectively.

\subsection{Main Results}
\subsubsection{Performance on Dialect ASR Task}
As illustrated in Table~\ref{tab:covost_asr_res}, the Whisper-based methods consistently outperform their Conformer-based counterparts. 
Specifically, the Conformer-based \textsc{FedAvg} and \textsc{FedLoRA} are both worse than \textsc{C-S2T}$_\textsc{Full}$. In contrast, the Whisper-based \textsc{FedLoRA} exhibits superior performance compared to both \textsc{C-S2T}$_\textsc{Full}$ and \textsc{FedAvg}. This improvement can be attributed to the fact that, with sufficient pre-training (Whisper employs 680k hours of speech data for pre-training), the lightweight LoRA component effectively overcomes catastrophic forgetting, enabling the preservation of pre-trained knowledge and the integration of client data. 
Unlike the full model interaction of \textsc{FedAvg}, the lightweight LoRA component mitigates the impact of model parameter conflicts arising from data heterogeneity to some extent. 
Moreover, considering that \textsc{FedLoRA} reduces the communication cost by 96.5$\%$ and 91.4$\%$ of \textsc{FedAvg} via lightweight LoRA module update and interaction, collectively \textsc{FedLoRA} is efficient and effective. Furthermore, \textsc{FedMem} enhances all FL methods, outperforms \textsc{C-S2T}$_\textsc{Full}$ by 0.61 WER on average in Whisper-based methods. This indicates that client-specific memorization retrieval can be effective in personalizing the global model to further overcome the data heterogeneity.

\subsubsection{Performance on Dialect ST Task.}
For the ST task, we observe similar trends as ASR task from Table~\ref{tab:covost_st_res}.
Specifically, in comparison to the \textsc{C-S2T}, the performance of \textsc{FedAvg} and \textsc{FedLoRA} are negatively affected by data heterogeneity.
\textsc{FedLoRA} effectively reduces the communication cost by 95.5\% of of that incurred by \textsc{FedAvg}. Furthermore, when \textsc{FedMem} is adopted, the average BLEU scores of \textsc{FedAvg}/\textsc{FedLoRA} are improved by 0.38/0.52, respectively. This demonstrates the efficiency and effectiveness of our approach on the ST task.

\subsubsection{Performance on Multi-domain ASR Task.}
To evaluate the scalability of our approach in a large-scale client setting, we conducted experiments using the multi-domain Gigaspeech ASR benchmark, which inherently exhibits a higher degree of data heterogeneity. As depicted in Table~\ref{tab:gigaspeech_asr_res}, we observed that \textsc{FedLoRA} significantly surpasses \textsc{FedAvg} by an average of 0.73 WER, while concurrently reducing the communication cost by 91.7\%. Moreover, when integrated with \textsc{FedMem}, the average performance of \textsc{FedLoRA} is notably superior to both \textsc{C-S2T}$_\textsc{Full}$ and \textsc{C-S2T}$_\textsc{LoRA}$, with improvements of 0.69/0.53 WER, respectively. These results demonstrate that our proposed method exhibits high scalability in large-scale, highly heterogeneous federated learning settings. 

\subsection{Generalization to Invisible Clients}
To evaluate the generalization ability to new coming clients, we conduct experiments on 6 "training-invisible" clients of multi-domain benchmark. 
In these experiments, \textsc{FedMem} directly utilize the global models \textsc{FedAvg} and  \textsc{FedLoRA} to establish a datastore and equip $k$NN for memorization retrieval during inference. 
The results, as shown in Table~\ref{tab:gigaspeech_unseen_res}, indicate that \textsc{FedLoRA} outperforms \textsc{C-S2T} and achieves similar performance to \textsc{FedAvg}, suggesting that \textsc{FedLoRA} can effectively generalize to new clients. Furthermore, compared to \textsc{FedAvg}/\textsc{FedLoRA}, \textsc{FedMem} achieves an average improvement of 0.65/0.67 WER scores, respectively. This implies that newly arriving clients can easily learn personalized systems after global training.
\begin{table}[h]
    \small
    \centering
    \caption{The ASR performance of different Whisper-based methods generalize to the invisible clients. ``Edu.", "Entmt." and ``NP" refer to the ``Education", ``Entertainment" and ``Nonprofits", respectively. }
    \label{tab:gigaspeech_unseen_res}
    \setlength{\tabcolsep}{1.25mm}{
    \scalebox{0.95}{
    \begin{tabular}{l|cccccc|cccccccccc}
        \toprule
         & \textbf{News} & \textbf{Edu.} & \textbf{Entmt.} & \textbf{NP} & \textbf{People} & \textbf{S\&T} & \textbf{Avg} \\
        \midrule
        \textsc{P-S2T}                  & 17.77  & 18.76 & 28.01  & 22.58  & 21.39  & 19.34  & 21.31  \\
        \midrule
        \textsc{C-S2T}$_\textsc{Full}$  & 6.01   & 6.06  & 8.20 & 7.48  & 6.09  & 6.68  & 6.75 \\
        \textsc{FedAvg}                 & 6.31   & 5.97  & 7.87 & 7.02  & 6.20  & 6.54  & 6.65 \\
        + \textsc{FedMem}               & 5.22   & \bluebold{4.91}  & \bluebold{7.13}  & 6.88 & \bluebold{5.72}  & 6.15  & \bluebold{6.00} \\
        \midrule
        \textsc{C-S2T}$_\textsc{LoRA}$  & 6.10   & 6.07  & 8.43  & 7.35  & 7.07  & 6.77  & 6.97 \\
        \textsc{FedLoRA}                & 5.93   & 6.06  & 9.07  & 7.13  & 6.25  & 6.52  & 6.87 \\
        + \textsc{FedMem}               & \bluebold{5.04}   & 5.34  & 8.35  & \bluebold{6.72}  & 5.77  & \bluebold{5.98}  & 6.20 \\
        \bottomrule
    \end{tabular}}}
\end{table}

\section{Discussions}
\par\noindent\ignorespacesafterend \textbf{Communication Cost.}
When interacting for $r$ rounds between a server and $|C|$ clients, the communication costs for \textsc{FedAvg} and \textsc{FedLoRA} are, respectively, 
$H_\theta *  |C| + H_\theta * |C| * r * 2$ and $H_\theta *  |C| + H_\delta * |C| * r * 2$,
where $H_\theta \ \text{and} \ H_\delta$ indicate the hard disk storage of the backbone model and the LoRA module. Typically, $H_\delta$ is much smaller than $H_\theta$. As the number of clients and rounds increase, \textsc{FedLoRA} requires a lower communication load compared to \textsc{FedAvg}. In our experiments, \textsc{FedLoRA} approach reduce the communication cost by least to $91.4\%$ compared to \textsc{FedAvg}, making it more scalable for large-scale settings.
\par\noindent\ignorespacesafterend \textbf{Inference Overhead.}
The inference overhead of $k$NN memorization retrieval has been discussed in previous work~\cite{Du2022NonParametricDA,Dai2023SimpleAS}, which may slightly reduce the speed. In this work, we investigate it on the CoVoST ASR benchmark based on Conformer. By setting the batch size and $k$ to 16 and 8, respectively, we find that the average inference speed of \textsc{FedMem} across four dialects is $92.9\%$ of that of the \textsc{C-S2T}. In practice, such a decline is acceptable, as it represents a trade-off between performance and processing time. We could also substitute it with other memorization-retrieval variants~\cite{Meng2021FastNN,Martins2022ChunkbasedNN,Dai2023SimpleAS} to achieve better inference speed, we leave it as future work. 

\section{Conclusions}
In this paper, we present the first personalized federated S2T framework. We first design \textsc{FedLoRA} to leverage a lightweight LoRA module to reduce communication overhead. Then \textsc{FedMem} employs client-specific memorization retrieval for personalization to address data heterogeneity. Experimental results on various FL scenarios demonstrate the effectiveness and efficiency of our framework.
In the future, we would like to integrate our methods with large S2T models (e.g., Whisper-Large) and explore more memorization retrieval methods.

\vspace{-2pt}
\section{Acknowledgements}
We thank the anonymous reviewers for helpful feedback on early versions of this work.
This work was supported by the grants from National Natural Science Foundation of China (No.62222213, 62072423) and the USTC Research Funds of the Double First-Class Initiative (No.YD2150002009). Zhirui and Tong are the corresponding authors.

\small
\bibliographystyle{IEEEbib}
\bibliography{refs}

\end{document}